\documentclass{article}

\usepackage[final]{corl_2020} 
\usepackage{amsmath}
\DeclareMathOperator{\atantwo}{atan2}

\usepackage{siunitx}

\usepackage{graphicx}
\usepackage{subcaption}
\usepackage{todonotes}

\usepackage{multirow}

\title{Obtaining Robust Control and Navigation Policies for Multi-Robot Navigation via Deep Reinforcement Learning}

\author{
  Chrstian~Jestel\(^{1,*}\), Hartmut~Surmann\(^2\), Jonas~Stenzel\(^1\),\\
  \textbf{Oliver~Urbann\(^1\), and Marius~Brehler\(^1\)}\\
  \(^{1}\)Automation and Embedded Systems, Fraunhofer IML, Germany \\
  \(^{2}\)Computer Science and Communication, Westf\"alische Hochschule Gelsenkirchen, Germany\\
  \(^*\)\texttt{christian.jestel@iml.fraunhofer.de}
}

\begin{document}
\maketitle


\begin{abstract}
Multi-robot navigation is a challenging task in which multiple robots must be coordinated simultaneously within dynamic environments. We apply Deep Reinforcement Learning (DRL) to learn a decentralized end-to-end policy which maps raw sensor data to the command velocities of the agent.
In order to enable the policy to generalize, the training is performed in different environments and scenarios. The trained policy is tested and evaluated in common multi-robot scenarios like switching a place, an intersection and a bottleneck situation. The learned policy allows the agent to recover from dead ends and to navigate through complex environments.
\end{abstract}



\section{Introduction}	
Multi-robot-navigation is one of the main challenges in mobile robotics. Multiple robots must be coordinated simultaneously to finish their task and have to navigate through a complex dynamic environment without causing collisions. One approach to enable the coordination of multi-robot navigation is prioritized planning, where robots plan their trajectories sequentially one after another. Prioritized planning algorithms tend to find a deadlock-free solution for route planning and centralized as well as decentralized planning solutions exist \cite{ppa_cap2015prioritized}. With a centralized approach all robots are coordinated by a single system, whereas navigation conflicts are resolved via communication between the robots in decentralized approaches. Prioritized path planning approaches tend to find solutions for scenarios with a high number of robots, while other approaches or reactive collision-avoidance algorithms like ORCA \cite{rvo_berg2008reciprocal} fail. However, the main drawback of centralized approaches is the bad scalability as the planning complexity increases drastically with the number of robots and the size and complexity of the environment \cite{stern2019multi}. Additionally, a reliable and synchronized communication between the centralized planner and all robots is essential. Decentralized approaches often rely on communication between robots in order to share state information (e.g. position and velocity~\cite{rvo_berg2008reciprocal} or reserved areas \cite{purwin2007path}) between all robots and thus require a reliable and flexible communication framework.

In recent years, advantages in the field of deep reinforcement learning (DRL) enabled algorithms to solve generic multi-robot navigation tasks in an end-to-end fashion.
End-to-end learning for robotic control refers to learning a model that transforms raw perception data directly to actuation commands for the robot.
This approach is completely decentralized and scales well when applied  to a larger number of robots. This is due to the deterministic planning time of the neural network that is independent of the number of robots. Furthermore, the approach does not rely on direct communication between the robots as every robot only uses its own local neural network as a controller.

In this paper, we compare different DRL training algorithms when applied to the multi-robot navigation problem. Beyond state-of-the-art approaches, we define and use new and more sophisticated reward functions.

\subsection{Related Work}
In this section, we provide an overview of different learning approaches that are used for DRL-based multi robot navigation.
Additionally, we give a brief comparison with other publications on DRL-based multi-robot navigation.

Mnih et al. made a breakthrough in the DRL domain with Deep Q Learning (DQN) by achieving excellent results in playing Atari games~\cite{dqn_mnih2015human}. Therefore, a lot of extensions have been implemented to DQN such as DDQN \cite{ddqn_van2016deep} and Rainbow~\cite{rainbow_hessel2018rainbow}.
Later on, the actor-critic model was successfully applied to several DRL problems~\cite{a3c_pmlr-v48-mniha16}. Various improvements to actor-critic model like PPO~\cite{ppo_schulman2017proximal} and A2C~\cite{a2c_acktr_NIPS2017_7112} were developed.
This set of training algorithms is compared in our DRL approach.

Baker et al. presented an emergent tool where a multi-agent system learned complex and emergent behavior using a PPO learning approach~\cite{hide_and_seek_baker2019emergent}.
A single agent can only observe a partial information about the environment but they learned to cooperate and evolve new strategies using a very simple reward scheme.
A kind of emergent behavior can also be seen in our approach as the robots learn to cooperate while navigating complex environments.
Fan et al. show a fully distributed reinforcement learning framework for mobile robot navigation~\cite{multi_agent_fan2020distributed}.
They show that a hybrid approach using a DRL robot controller combined with a traditional PID controller can overcome situations where the DRL controller fails. The learned policy is successfully transferred to different physical robots and is able to navigate in the real world. Compared to their approach we use a more sophisticated reward scheme (see section~\ref{sec:reward_function}) that may make the use of an additional PID controller obsolete. Furthermore, they do not evaluate the learned policies in completely unknown environments. Lin et al. present an approach where the robots know each other's position explicitly and where a communication between the robots is guaranteed \cite{lin2020connectivity}. As proposed in the conclusion of \cite{lin2020connectivity}, our approach is decentralized and a robot does not explicitly know the positions of the other robots. Hence, there is no necessity for explicit data communication between the robots.
In~\cite{gromniak2019deep} a similar approach as ours is shown, but it is limited to less complex environments and only one robot is navigating through the environment. 


\subsection{Contribution}
In this paper, we propose reward functions to be applied to DDQN, A2C and PPO training algorithms.
Those are used to solve the multi-robot navigation problem and compared with each other.
Furthermore, we introduce some multi-robot scenarios for the training of the system.
Accompanied by our parametrization, that can show that the learning behavior is comparable to that of a human, e.g. zip merging when driving a car.

\subsection{Overview}
This paper is structured as follows.
First, we explain the reward function, followed by a description of the architecture of the neural network.
In section~\ref{sec:experiments}, we first introduce the training algorithms and the environments used within the evaluation. Thereafter, the training algorithms are compared against each other and the best is used for the final training and evaluation. Finally, a conclusion is given in section~\ref{sec:conclusion}.


\section{Approach}
\label{sec:approach}

The multi-robot-navigation problem is defined as a nonholonomic differential driven robot which navigates in a plane environment to reach a given goal.
For the training multiple, different environments are used. 
Each environment contains up to $N$ robots as decision-making agents. 
All robots are represented by a circle shape, all with the same radius.
The robots are equipped with a 2D-laser scanner with a range of \SI{270}{\degree} and a resolution of \SI{0.25}{\degree}. 
Noise is added to the data of the 2D-laser scanner with a variance of \SI{+- 0.04}{\meter}.
Laser scanners are not only the main source of perception in industrial environments, in fact they are the only approved safety sensor for autonomous guided vehicles.
Thus, end-to-end control and navigation policies based on raw data of laser scanners that are efficient to compute are desirable.

Each agent operates independently from each other and no direct communication between the agents takes place.
Nevertheless, the robots communicate indirectly via the observation of the other robots.
The reward function is calculated for each agent and one episode which consists of multiple training steps, executed till a stop condition is reached. All results are collected to train a single neural network. This network is applied on all agents within the next episode. In the following, the approach is illustrated for a single agent.

For each timestep $t$ the agent gets the observation $o^t$ and predicts an action $a^t$, which is a velocity command to reach the given goal $g^t$. The agent cannot fully observe the environment with its sensors. The observation is split into four parts $o^t = [o_l, o_g, o_d, o_v]$, where $o_l$ is the measured laser distance of the 2D-Lidar sensor, $o_g$ is the relative orientation to the goal as a direction vector in the robot coordinate system, $o_d$ is the Euclidean distance to the goal and $o_v$ represents the current linear and angular velocity of the robot. For every agent the action $a^t$ is predicted independently, incorporating the shared policy
\begin{align}
 a^t \sim \pi_\theta (a^t|o^t)
\end{align}

where $\theta$ is the policy parameter.
The agent's action $a^t\mathrel{\widehat{=}}v^t$ is defined by its differential driven actuator which consist of the linear velocity $v_{\textrm{lin}}=[0.0, 0.6]\,\si{\meter\per\second}$ and angular velocity $v_{\textrm{ang}}=[-1.5, 1.5]\,\si{\radian\per\second}$. The agent is not allowed to drive backwards because the 2D-laser-scanner's range of \SI{270}{\degree} implies a blind spot in the agent's back.

\subsection{Reward Function}
\label{sec:reward_function}
The reward function is defined as
\begin{align}
\label{eq:rew}
r^t = \begin{cases}
       \vartheta_{\textrm{goal}} &\textrm{goal reached}\\
       \vartheta_{\textrm{collision}} &\textrm{collision}\\
       0 & \textrm{timeout}\\
       r_{\textrm{dist}}^t + r_{\textrm{ori}}^t + r_{\textrm{sd}}^t + r_{\textrm{mld}}^t + r_{\textrm{wig}}^t &\textrm{else}
      \end{cases}
\end{align}
where $\vartheta_{\textrm{goal}}$ is the reward for reaching the goal and $\vartheta_{\textrm{coll}}$ the negative reward for a collision prior to reaching the goal. No reward is given if a timeout occurs. If only these three cases are considered, it is hard for a continuous optimization to find better solutions in small steps as the reward does not change. Thus, we define an additional case containing the sum of fine-grained dense rewards. In the following, we describe all summands of this case and explain $\vartheta_{\textrm{coll}}$ and $\vartheta_{\textrm{goal}}$ afterwards.

\subsubsection{Distance to Goal}
To encourage the agent to move into the direction of the goal, the ``distance to goal'' reward $r_{\textrm{dist}}$ is based on the change of the distance to the goal $\Delta d$ from the previous timestep $t-1$ to the current timestep $t$:
\begin{align}
\Delta d = d_{\textrm{goal}}^{t-1} - d_{\textrm{goal}}^t
\end{align}
Here, $d_{\textrm{goal}}^t = ||\vec{g^t}||$ is the distance to the goal containing the vector from current position to goal $\vec{g^t}  =  g - p^t$. The position of the agent at time $t$ is given by $p^t$ and $g$ is the goal for agent. 

However, the agent should also be encouraged to increase the distance to the goal in case it has to leave a dead end or to drive around an obstacle. Thus, $r_{\textrm{dist}}$ is split into two parts, allowing to scale down the negative reward or even set it to zero if required:
\begin{align}
r_{\textrm{dist}} = \begin{cases}
                     \Delta d \cdot d_{\textrm{neg}} & \Delta d < 0 \\
                     \Delta d \cdot d_{\textrm{pos}} & \textrm{else} \\
                    \end{cases}
\end{align}
Here, $d_{\textrm{pos}}$ and $d_{\textrm{neg}}$ are the scaling factors for a positive and a negative reward, respectively.

\subsubsection{Orientation to Goal}
The partial reward $r_{\textrm{ori}}$ is defined by the orientation of the agent to the goal. First, the current angle~$\alpha$ of the agent to the goal is calculated as:
\begin{align}
\alpha = |\atantwo(\Vert\vec{o^t} \times\vec{g^t}\,\Vert, \vec{o^t}\cdot\vec{g^t})|
\end{align}
Here, $\alpha \le \SI{90}{\degree}$ is desired. In this case the robot is decreasing the orientation to the goal.
Therefore, the value $\alpha_{\textrm{norm}}$ is defined as:
\begin{align}
\alpha_{\textrm{norm}} = 1 - \frac{2\alpha}{\pi}
\end{align}
This value is positive if the angle is within the desired range and negative otherwise. By adding the scaling factors $\alpha_{\textrm{pos}}$ and $\alpha_{\textrm{neg}}$ we get:
\begin{align}
r_{\textrm{ori}} = \begin{cases}
                    \alpha_{\textrm{norm}} \cdot \alpha_{\textrm{neg}} & \alpha_{\textrm{norm}} < 0\\
                    \alpha_{\textrm{norm}} \cdot \alpha_{\textrm{pos}} & \textrm{else}
                   \end{cases}
\end{align}
Compared to the ``distance to goal'' reward function, the ``orientation to goal'' reward should be weighted low. Otherwise the agent orients itself towards goal, but does not moves further in case of an obstacle.

\subsubsection{Shortest Distance to Goal}
The partial reward function ``shortest distance to goal'' $r_{\textrm{sd}}$ is comparable to the ``distance to goal'' function $r_{\textrm{dist}}$. However, a positive reward is only given, if the robot can further reduce the distance to goal compared to the situation at the beginning of the current episode. This can avoid situations in which the agent is trapped moving only back and forth, whereas $r_{\textrm{dist}}$ would reward this.

A variable $l$ is therefore initialized as $l=d_{\textrm{goal}}^0$ and updated with $d_{\textrm{goal}}^t$ if $ d_{\textrm{goal}}^t < l$. With $l_{\textrm{pos}}$ as scaling factor, we define:
\begin{align}
r_{\textrm{sd}}^t = \begin{cases}
                      (l-d_{\textrm{goal}}) \cdot l_{\textrm{pos}} & d_{\textrm{goal}} < l \\
                      0 & \textrm{else}
                     \end{cases}
\end{align}
As one can see, the robot is only rewarded if the distance is below the absolute minimum within this episode up to $t$ but does not get a negative reward to drive out of dead ends.

\subsubsection{Minimum Laser Distance}
Collisions should obviously be avoided and thus the reward function needs to take a negative reward into account for that case. Since we assume that the robot is equipped with a laser scanner, the probability for a collision can be minimized by giving a negative reward for approaching an obstacle below a given minimum distance. We define the minimum distance as the size of the robot $r_{\textrm{collision}}$ (assuming that it is a round shaped robot) and $l_{\textrm{laser}}$. The latter is a manual chosen minimum distance of the agent's case to an obstacle. It is directly measurable using the laser scanner. This minimum can be used to define the partial reward function
\begin{align}
r_{\textrm{mld}}^t \! = \! \begin{cases}
                      (r_{\textrm{collision}} \! + \! l_{\textrm{laser}} \! - \! l_{\textrm{min}}) \! \cdot \! (-l_{\textrm{neg}}) & l_{\textrm{min}} \! < \! r_{\textrm{collision}} \! + \! l_{\textrm{laser}}\\
                      0 & \textrm{else}
                     \end{cases}\,,
\end{align}
where $l_{\textrm{neg}}$ is a manually chosen scaling factor and $l_{\textrm{min}}$ is the measured distance of the laser beam currently measuring the lowest distance of all beams.

\subsubsection{Limitation of Redirection}
With $r_{\textrm{dist}}$ and $r_{\textrm{sd}}$ we reward the agent to drive to the goal while avoiding obstacles. In order to do so, a change of direction is indirectly rewarded. However, driving straight forward is in most cases the fastest movement. Thus, a change of direction should get a negative reward $r_{\textrm{wig}}$ if it happens too often.

Let $\Delta \omega_{\textmd{wig}}^t$ be the change of direction from $t-1$ to $t$ and $\omega_{\textrm{dir}}$ a threshold. We then define
\begin{align}
F^t = \begin{cases}
       \textrm{left}& \Delta \omega_{\textmd{wig}}^t > \omega_{\textrm{dir}} \\
       \textrm{right}& \Delta \omega_{\textmd{wig}}^t < \omega_{\textrm{dir}} \\
       \textrm{straight}& \textrm{else}
      \end{cases}\,.
\end{align}
We can now define a change of direction $R^t$ if $F^t$ changes as
\begin{align} 
R^t = \begin{cases}
            1 & \big[\; (F^t = \textrm{left} \wedge F^{t-1} = \textrm{right}) \\
              & \vee (F^t = \textrm{right} \wedge F^{t-1} = \textrm{left}) \big] \\
            0 & \textrm{else}
            \end{cases}\,.
\end{align}
Limiting to \(N\) changes during a period \(T\) finally leads to
\begin{align}
r_{\textrm{wig}}^t = \begin{cases}
       \frac{\omega_{\textmd{neg}}}{T} \sum^{T}_{t=0} R^t& N<\sum^{T}_{t=0} R^t\\
       0 & \textrm{else}
      \end{cases}\,.
\end{align}

\subsubsection{Final States}
The sum of the afore presented dense rewards, see \eqref{eq:rew}, is applied if no final state was reached yet (goal not reached, no collision and still driving to goal). Final states are treated separately. The positive reward $\vartheta_{\textrm{goal}}$ for reaching the goal is constant. The negative reward depends on the type of collision:
\begin{align}
\label{eq:finstates}
\vartheta_{\textrm{collision}}^t = \begin{cases}
                                    - c_{\textrm{world}}& \textrm{collision with world} \\
                                    - c_{\textrm{robot}} & \textrm{collision with robot}
                                   \end{cases}
\end{align}
The values of $\vartheta_{\textrm{goal}}$ and the constants \(c\) in \eqref{eq:finstates} should be chosen significantly higher than the dense rewards. This is because reaching the goal without collision has the highest priority and the dense rewards are only required to give a meaningful feedback for small training steps.


\subsection{Network Architecture}

The neuronal network consists of 7 hidden layers. 
The observation, which is the given input $o^t$, is directly mapped to the agent's velocity $v^t$ as output. As in \cite{dqn_mnih2015human}, the last four observations are used as input. Hence, the agent can detect the motion of the other agents through the change of laser beams and also knows its own travelled path for a short duration, here for a second. The input of the laser scanner $o_l$ is connected to a 1D-convolution with 16 filters, a kernel size = 7, and a stride~=~3. This is followed by a second 1D-convolution layer with 32 filters, a kernel size = 5, and a stride = 2. Subsequently a fully connect layer with 256 units is used. The other three inputs $o_d, o_g$ and $ o_d$ are connected each with its own fully connected layer, where the layer of $o_g$  has 32 units, $o_d$ 16 units and $o_v$ 32 units. Finally, all four fully connected layers are merged by a fully connected layer with 384 units. For all seven hidden layers the ReLU activation function is used.

The output varies between the discrete and continuous action space. To compare all three algorithms DDQN, A2C and PPO, the agent has ten discrete actions available.
For the continuous action space,~$a_t$ is sampled from mean of a Gaussian distribution with variable standard deviations.


\section{Experiments}
\label{sec:experiments}
In this section, the training setup as well the hyperparameter for the training algorithms and the selected environment are described. Furthermore, the three training algorithms are compared. Afterwards the result of the known worlds, which are used for the training, are evaluated. Next the agent's performance is evaluated in unknown worlds. In one setup the agent is only trained in three environments, but its behavior is evaluated in two unknown environments. The other setup shows the agent's behavior after the training with the multi-robot scenario in an unknown world, which contains elements from the training environments.

\subsection{Training}
\label{sec:training}
For the training, the robust and efficient training algorithm Proximal Policy Optimization (PPO) \cite{ppo_schulman2017proximal} is used. We further compare PPO with the vanilla version of the actor-critic model A2C. The implementation is similar to the PPO and the A2C of the OpenAI baseline \cite{baselines}. For every agent state $s_t$ in an environment, the policy $\pi_\theta$ is executed.  Subsequently, the trajectory is sampled from all agent with generalized advantage estimation algorithm \cite{gae_Schulman2016High}. In PPO the sampled training data are shuffled and split into multiply minibatches to update the policy. The policy in A2C is updated with all sampled training data at once. The hyperparameters for the learning algorithms are given in Table~\ref{tab:hyperparameters-alg}, the ones for the reward function are given in Table~\ref{tab:hyperparameters-func}.

\begin{table}[h!]
	\centering
	\caption{Hyperparameters of the learning algorithms.}
	\label{tab:hyperparameters-alg}
	{\renewcommand{\arraystretch}{1.25}
		\begin{tabular}{|l|l||l|l|}
			\hline
			\multicolumn{2}{|c||}{\textbf{DDQN}} & \multicolumn{2}{c|}{\textbf{A2C and PPO}}\\
			\hline
            Optimizer                    & Adam                       & Optimizer                  & Adam\\
            Learning rate                & $0.00005$                  & Learning rate              & $0.0003$\\
			Discount $\gamma$            & $0.95$                     & Discount $\gamma$          & $0.99$\\ 
            Batch size                   & $64$                       & $T_{\textrm{max}}$         & $64$\\
            Update $Q_{\textrm{target}}$ & $250$                      & GAE parameter $\lambda$    & $0.95$\\
                                         &                            & PPO clipping $\varepsilon$ & $0.2$\\
                                         &                            & PPO minibatch size         & $4096$\\
			\hline
	\end{tabular}}
\end{table}

\begin{table}[h!]
	\centering
	\caption{Hyperparameters of the reward function.}
	\label{tab:hyperparameters-func}
		{\renewcommand{\arraystretch}{1.25}
		\begin{tabular}{|l|l|l|l|l|}
			\hline
            $\vartheta_{\textrm{goal}}$ & $c_{\textrm{world}}$ & $c_{\textrm{robot}}$ & $d_{\textrm{pos}}$ & $d_{\textrm{neg}}$ \\
            \hline
            $1.0$ & $0.75$ & $1.0$ & $0.01$ & $0.002$ \\
			\hline
	        $\alpha_{\textrm{pos}}$ & $\alpha_{\textrm{neg}}$ & $l_{\textrm{pos}}$ & $l_{\textrm{neg}}$ & $\omega_{\textmd{neg}}$  \\
			\hline
			$0.001$ & $0.0002$ & $0.05$ & $0.01$ & $0.01$ \\
			\hline
	\end{tabular}}
\end{table}

\begin{figure}[h!]
	\centering
	\begin{subfigure}[c]{0.48\columnwidth}
		\centering
		\includegraphics[width=\textwidth]{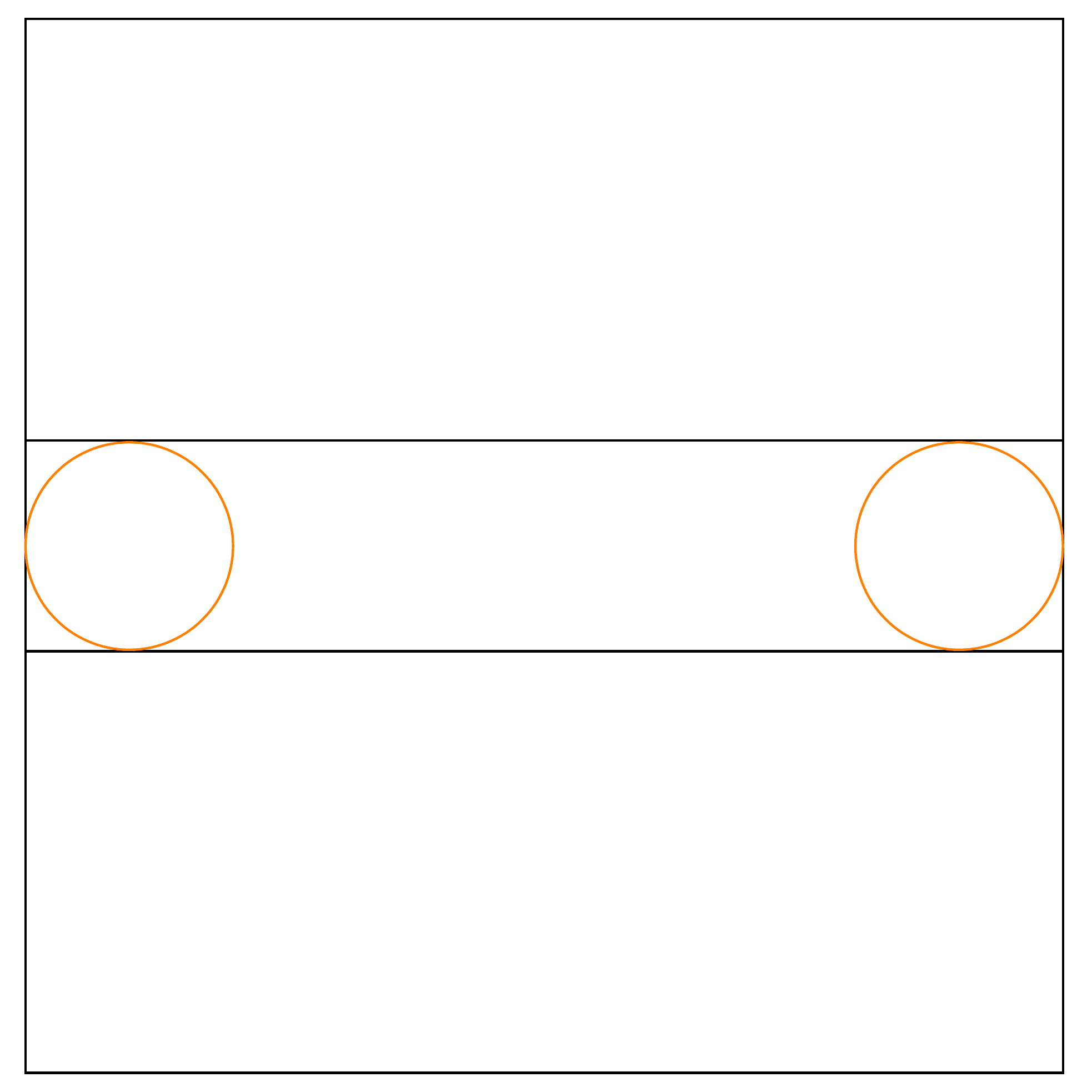}
		\caption{Tube.}
	\end{subfigure}
	\begin{subfigure}[c]{0.48\columnwidth}
		\centering
		\includegraphics[width=\textwidth]{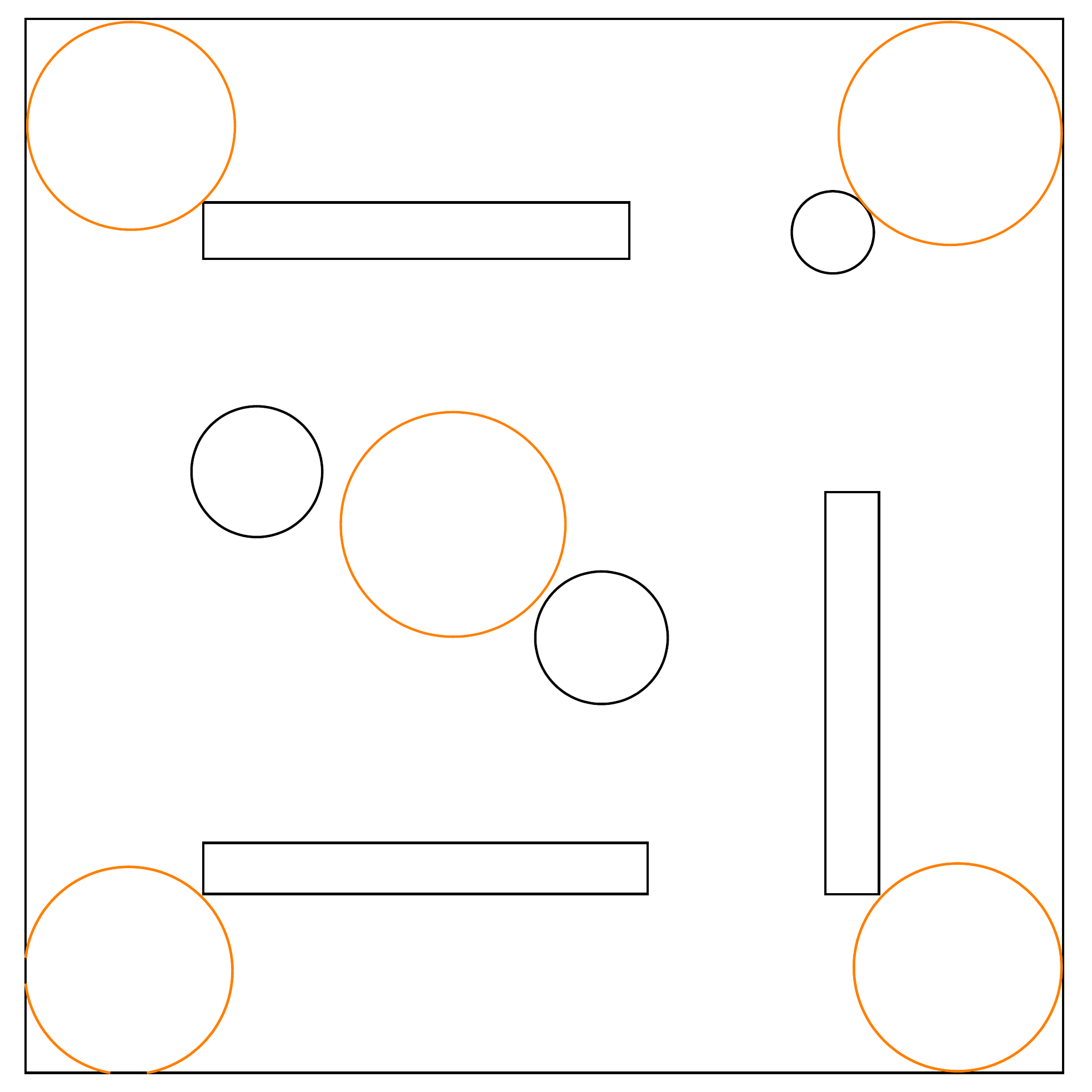}
		\caption{Room.}
	\end{subfigure}

	\begin{subfigure}[c]{0.48\columnwidth}
		\centering
		\includegraphics[width=\textwidth]{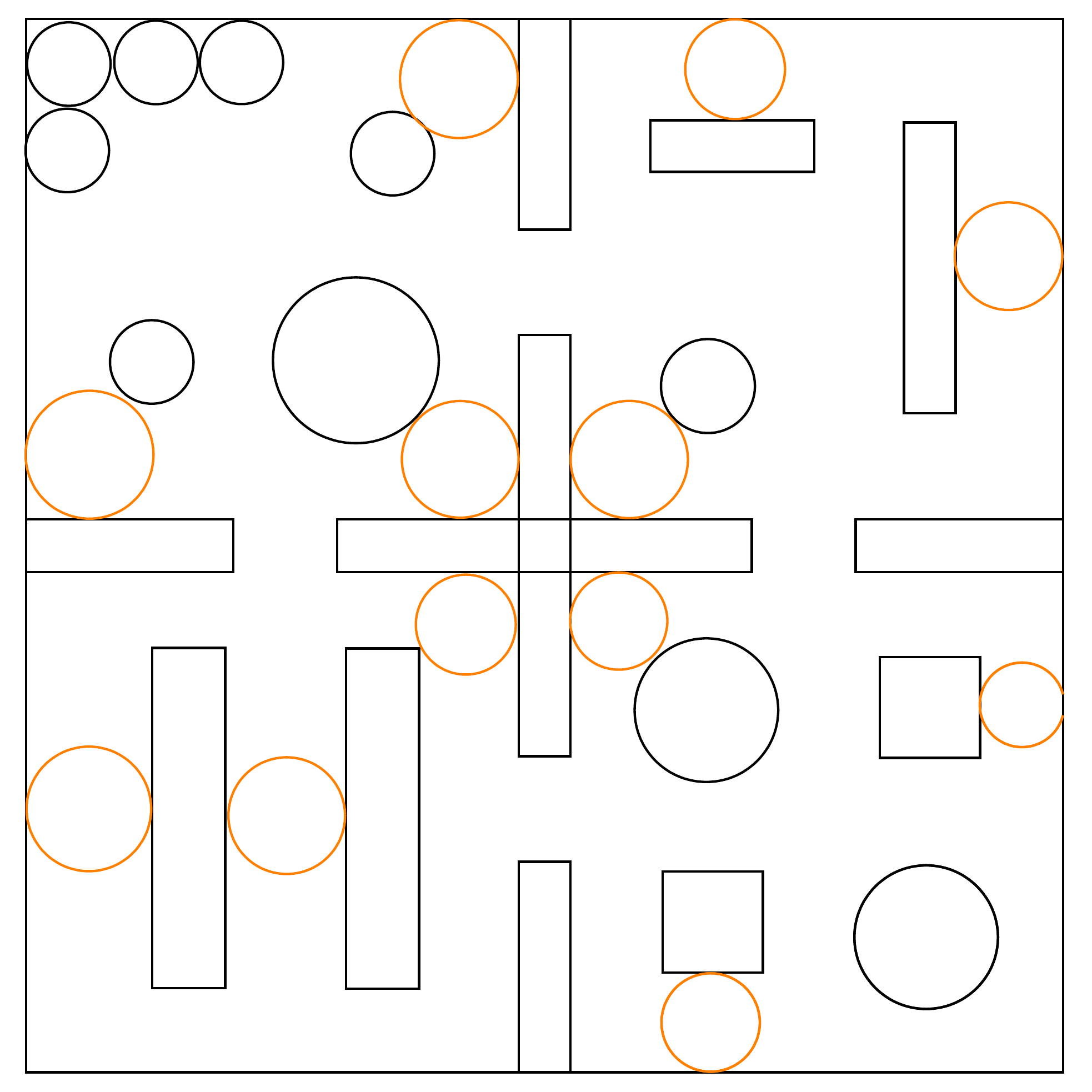}
		\caption{Four rooms.}
	\end{subfigure}
	\begin{subfigure}[c]{0.48\columnwidth}
		\centering
		\includegraphics[width=\textwidth]{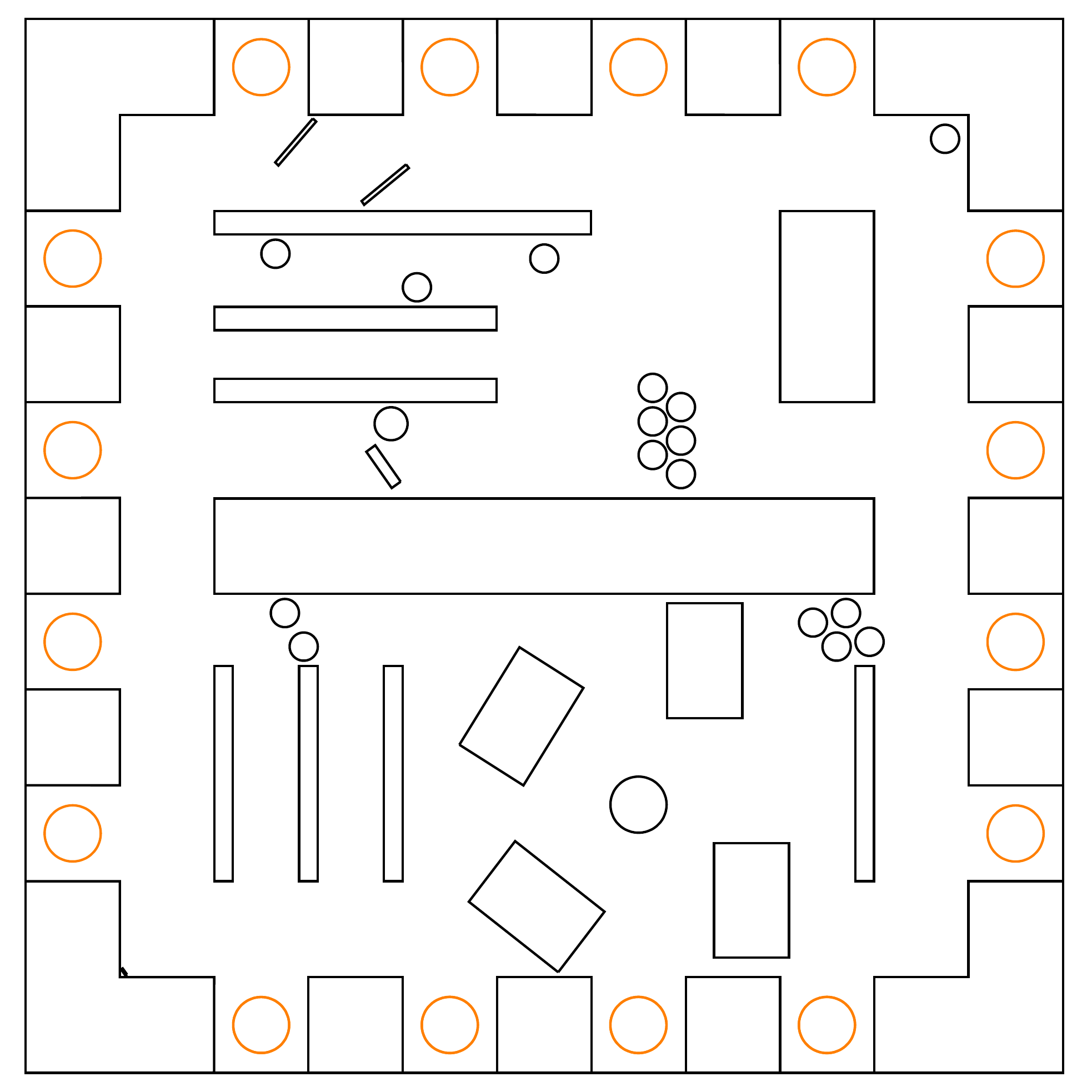}
		\caption{Hall.}
	\end{subfigure}

	\begin{subfigure}[c]{0.48\columnwidth}
		\centering
		\includegraphics[width=\textwidth]{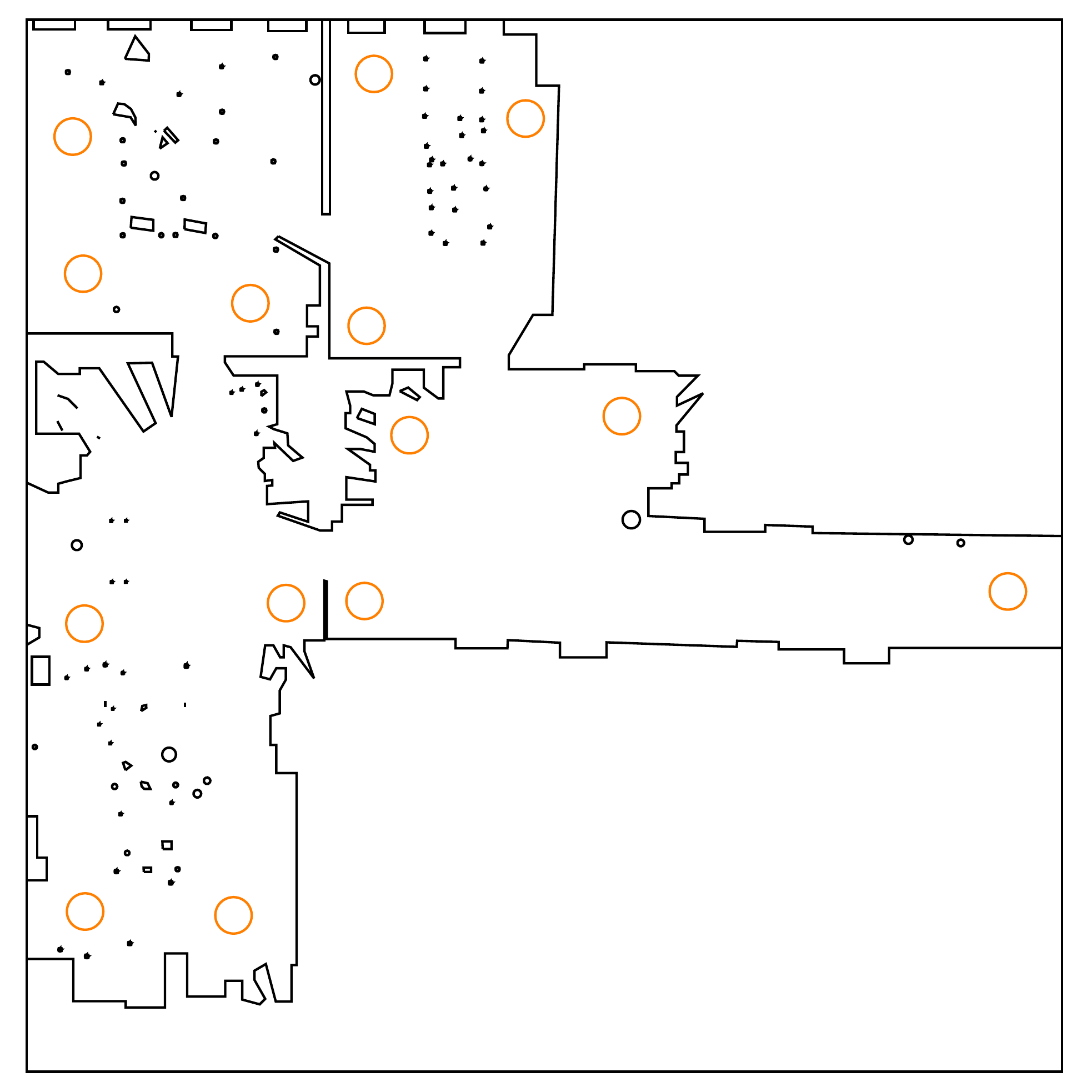}
		\caption{Roblab.}
	\end{subfigure}
	\begin{subfigure}[c]{0.48\columnwidth}
		\centering
		\includegraphics[width=\textwidth]{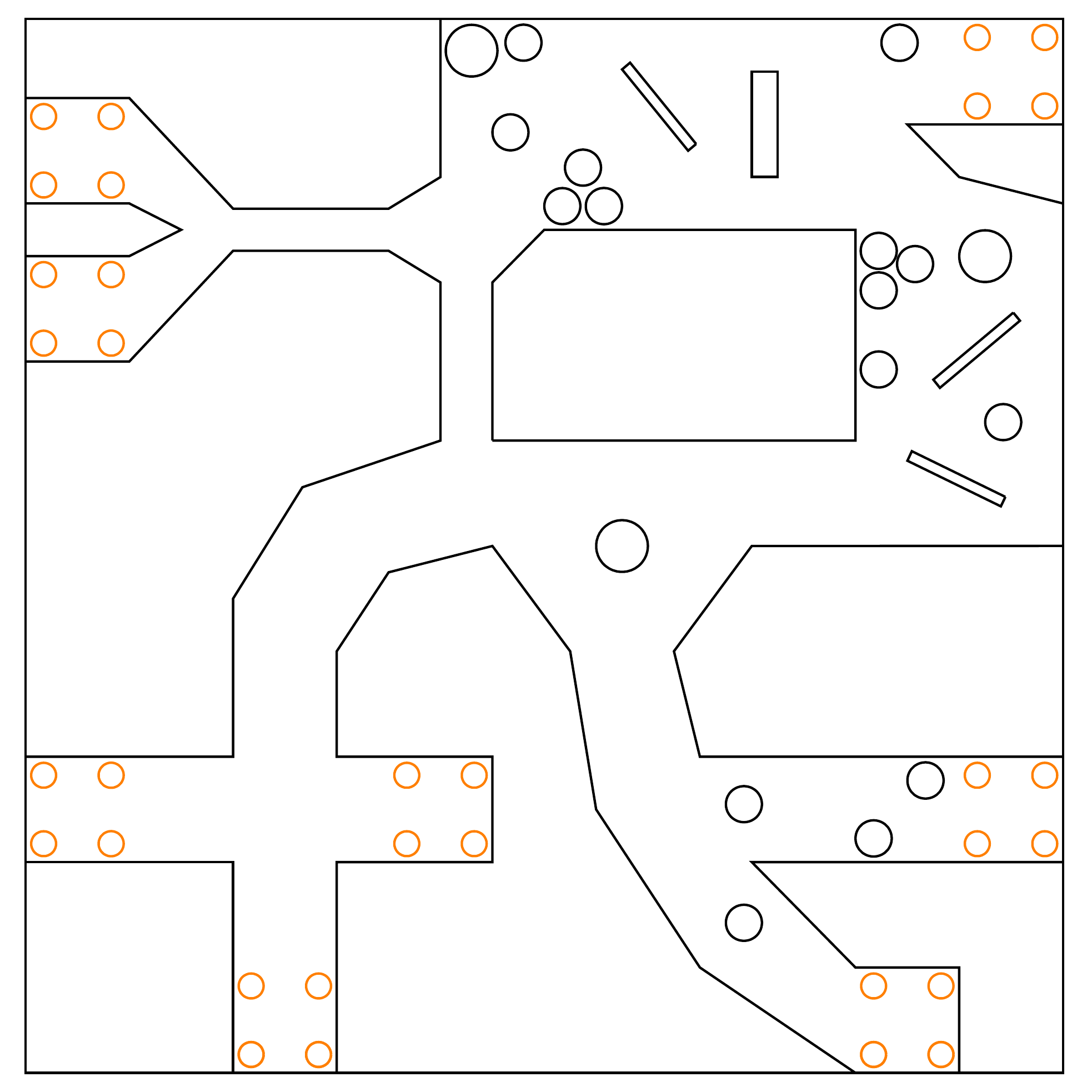}
		\caption{Multi.}
		\label{fig:environment_multi}
	\end{subfigure}
	\caption{The used environment for multi-robot navigation problem. The obstacles are marked in black and the nodes for a possible start and goal are orange.}
	\label{fig:environments}
\end{figure}

We have chosen various environments with different challenges. Further, we designed four environments for typical multi-robot scenarios. Fig.~\ref{fig:environments} and Fig.~\ref{fig:eval_know_worlds} depict the environments. Additionally, and not depicted in the figures, a scenario in which eight agents must pass a small bottleneck is trained and evaluated. The number of agents used in the different environments is given in Table~\ref{tab:known_worlds}.
%
%

\begin{figure}[h!]
 \centering
 \includegraphics[width=\columnwidth]{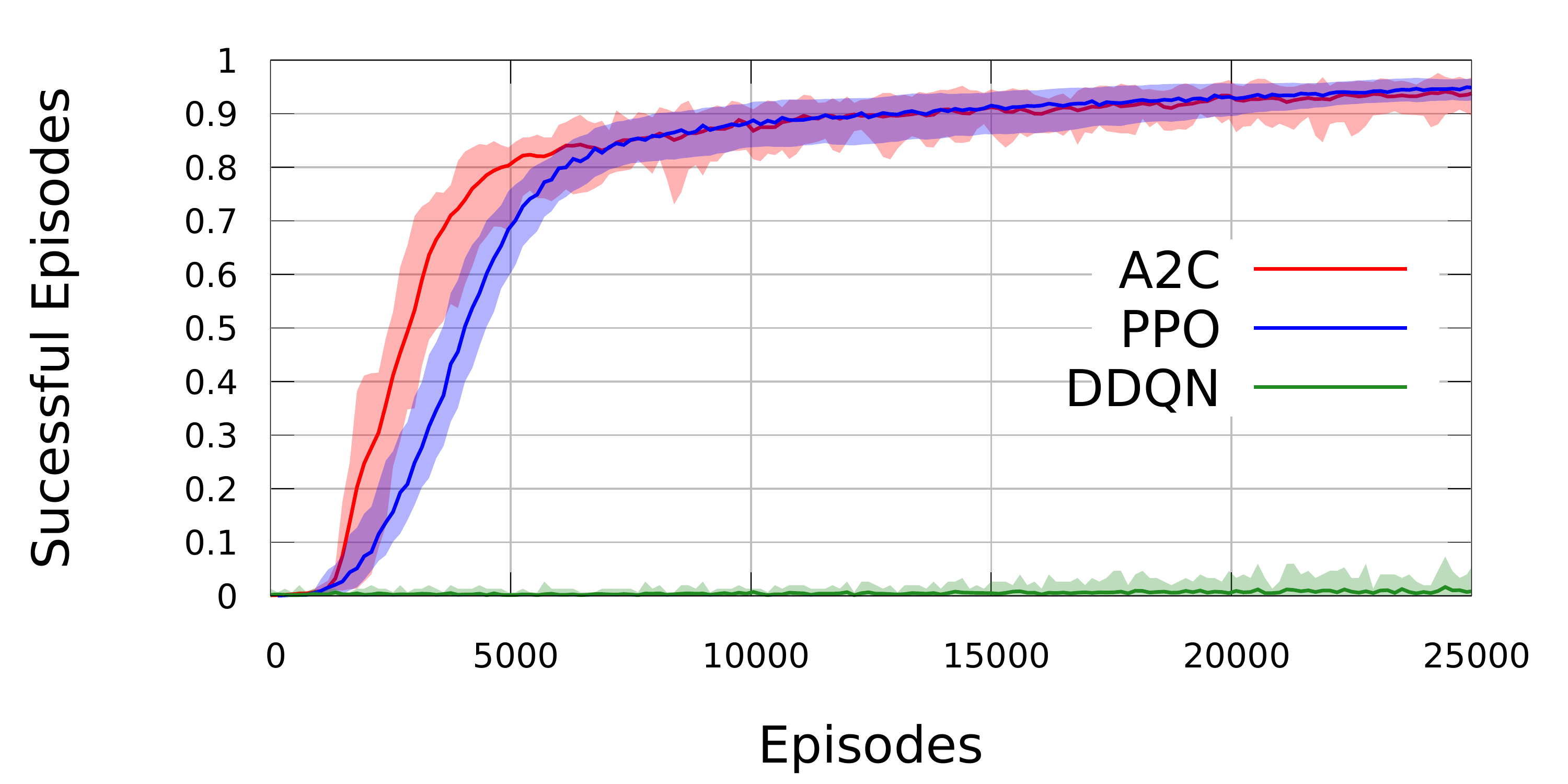}
 \caption{Comparison of the training algorithms: A2C (red), PPO (blue) and DDQN (green).}
 \label{fig:comparision_algos}
\end{figure}

In a first step, we evaluated which of the training algorithms DDQN, A2C and PPO is suited best for the multi-robot navigation reinforcement learning problem. Here, the three environments ``tube'', ``room'', and ``four rooms'' are utilized. These are simple to learn and thus lead to a fast convergence.  Each algorithm is executed ten times until 25.000 episodes are reached. As shown in Fig.~\ref{fig:comparision_algos}, DDQN does not converge even with more training episodes than shown in Fig.~\ref{fig:comparision_algos}. A2C and PPO are successfully convergent. PPO is converging slower than A2C, but PPO is more stable.

Since PPO works best of the three algorithms in the discrete action space, and since it is further shown in \cite{ppo_schulman2017proximal} that PPO works quite well for the continuous action space, PPO is chosen for all other experiments in the continuous action space.

\subsection{Evaluation on Known Worlds}
\begin{table}[h!]
 \centering
 \caption{Results of evaluation in known worlds.}
 \label{tab:known_worlds}
 {\renewcommand{\arraystretch}{1.25}
 \begin{tabular}{|l|l|l|l|l|}
 \hline
 \textbf{Environment} & \textbf{Num. agents} & \textbf{Reached goal} & \textbf{Timeout} & \textbf{Collision} \\
 \hline
 tube & 2          & \SI{100.00}{\percent} & \SI{0.00}{\percent} & \SI{0.00}{\percent} \\
 room & 5          & \SI{100.00}{\percent} & \SI{0.00}{\percent} & \SI{0.00}{\percent} \\
 four rooms & 12   & \SI{99.69}{\percent}  & \SI{0.03}{\percent} & \SI{0.28}{\percent} \\
 hall & 16         & \SI{98.38}{\percent}  & \SI{1.39}{\percent} & \SI{0.23}{\percent} \\
 roblab & 14       & \SI{98.28}{\percent}  & \SI{0.01}{\percent} & \SI{1.71}{\percent} \\
 swap & 16         & \SI{100.00}{\percent} & \SI{0.00}{\percent} & \SI{0.00}{\percent} \\
 intersection & 8  & \SI{100.00}{\percent} & \SI{0.00}{\percent} & \SI{0.00}{\percent} \\
 intersection & 16 & \SI{99.99}{\percent}  & \SI{0.00}{\percent} & \SI{0.01}{\percent} \\
 bottleneck & 4    & \SI{100.00}{\percent} & \SI{0.00}{\percent} & \SI{0.00}{\percent} \\
 bottleneck & 8    & \SI{99.44}{\percent}  & \SI{0.39}{\percent} & \SI{0.17}{\percent} \\
 constriction & 8  & \SI{99.89}{\percent}  & \SI{0.04}{\percent} & \SI{0.07}{\percent} \\
 \hline
 \end{tabular}}
\end{table}
All worlds listed in Table~\ref{tab:known_worlds} are trained in parallel. For each update of the policy the merged training data of all those worlds are used. The training is terminated when an average success rate of \SI{99.5}{\percent} is reached, which was the case after 2.500.612 episodes. In most environments the agent reached a success rate of \SI{100}{\percent}. Lower success rates were reached in the environments ``hall'' and ``roblab'', which is explained more detailed in the section \ref{sec:unknown_worlds}.

\begin{figure}[h!]
	\centering
	\begin{subfigure}[c]{0.48\columnwidth}
		\centering
		\includegraphics[width=\textwidth]{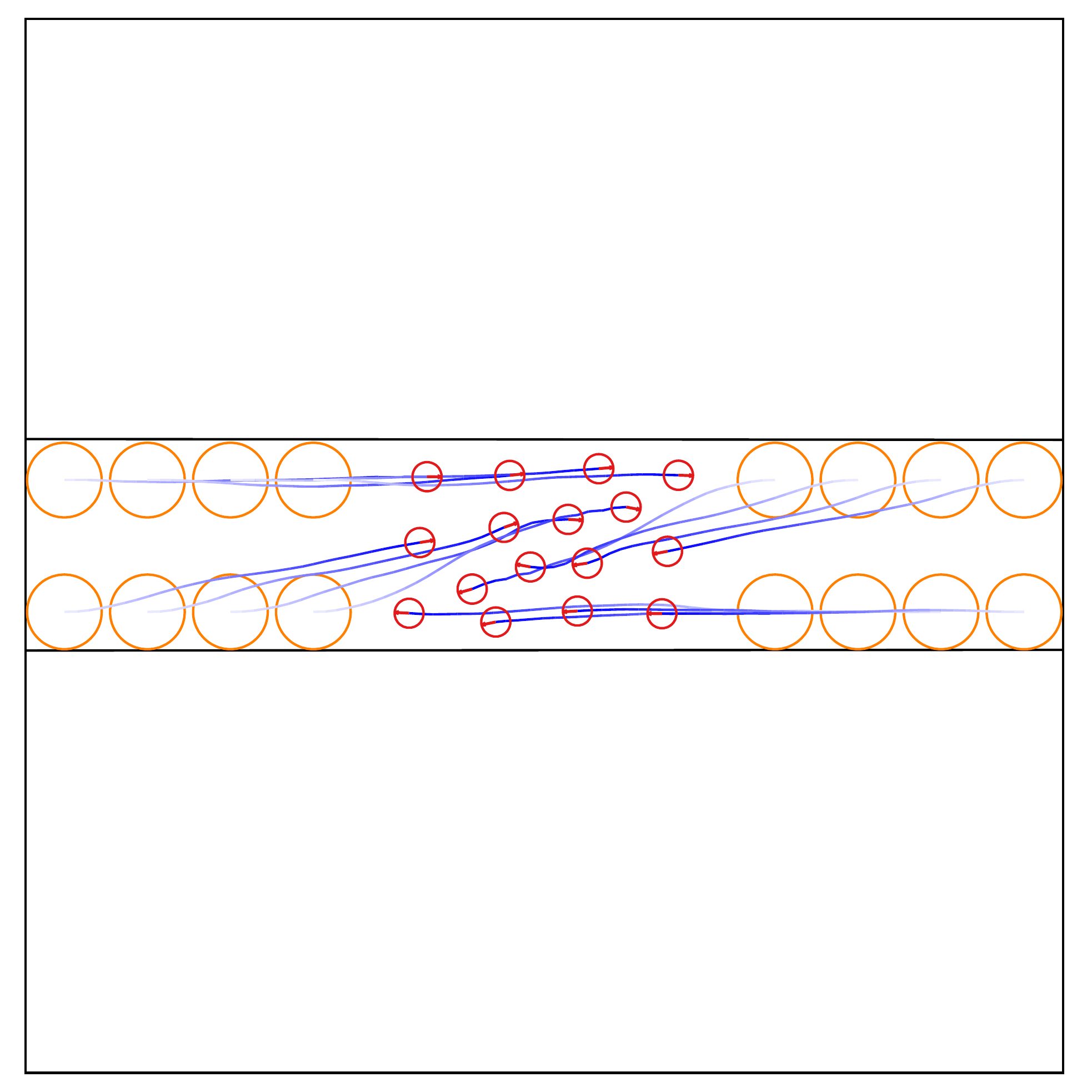}
		\caption{Swap.}
	\end{subfigure}
	\begin{subfigure}[c]{0.48\columnwidth}
		\centering
		\includegraphics[width=\textwidth]{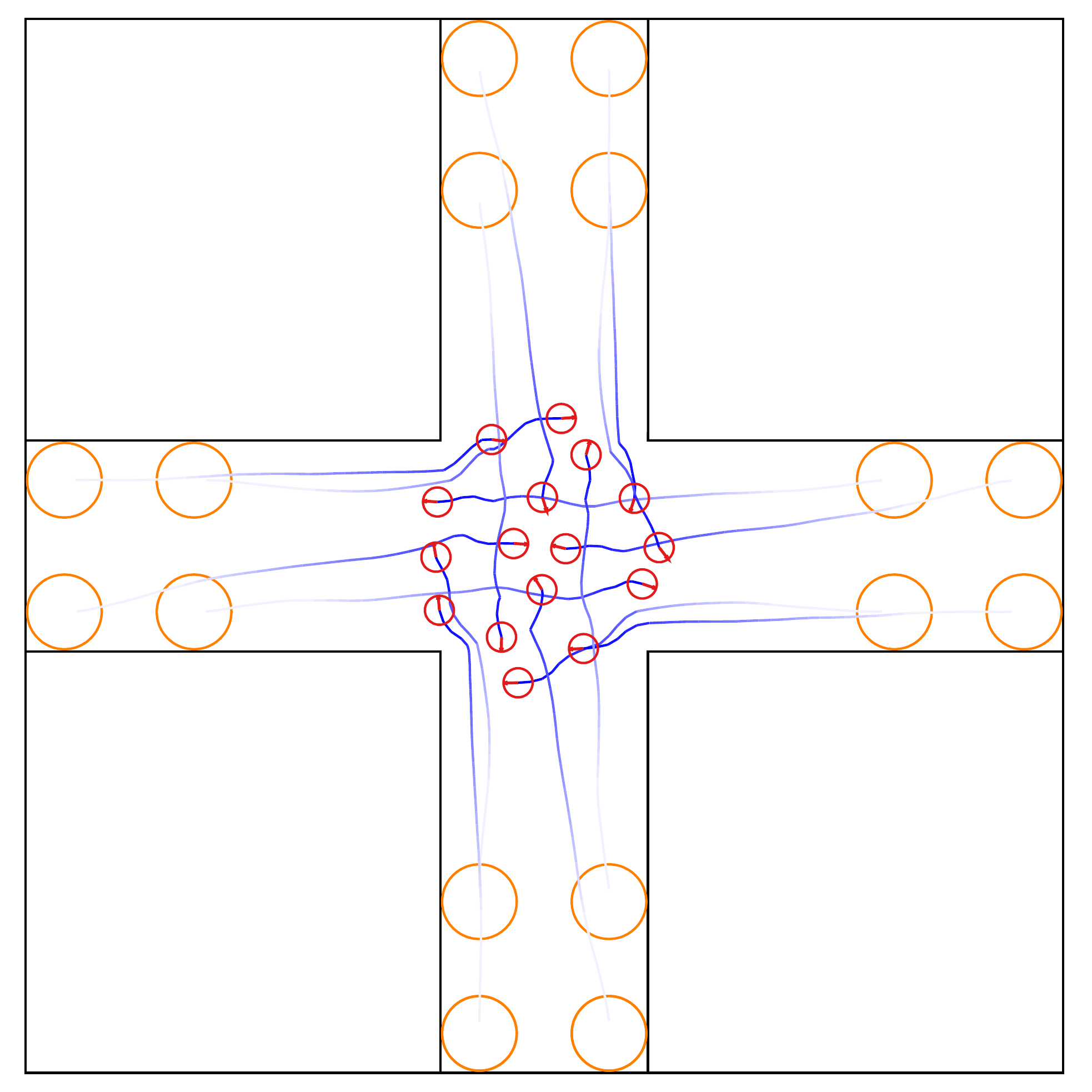}
		\caption{Intersection.}
	\end{subfigure}
	
	\begin{subfigure}[c]{0.48\columnwidth}
		\centering
		\includegraphics[width=\textwidth]{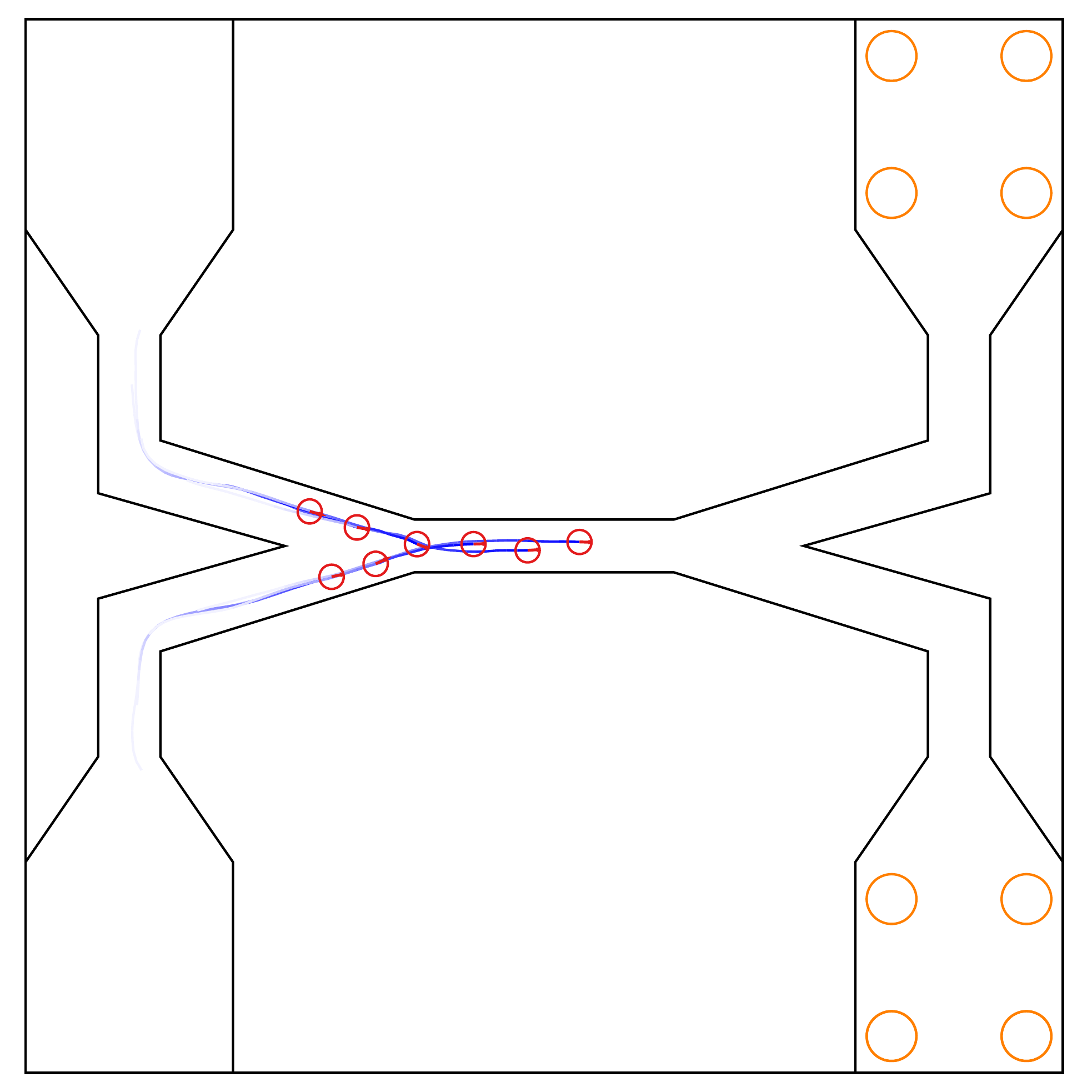}
		\caption{Constriction.}
	\end{subfigure}
	\begin{subfigure}[c]{0.48\columnwidth}
		\centering
		\includegraphics[width=\textwidth]{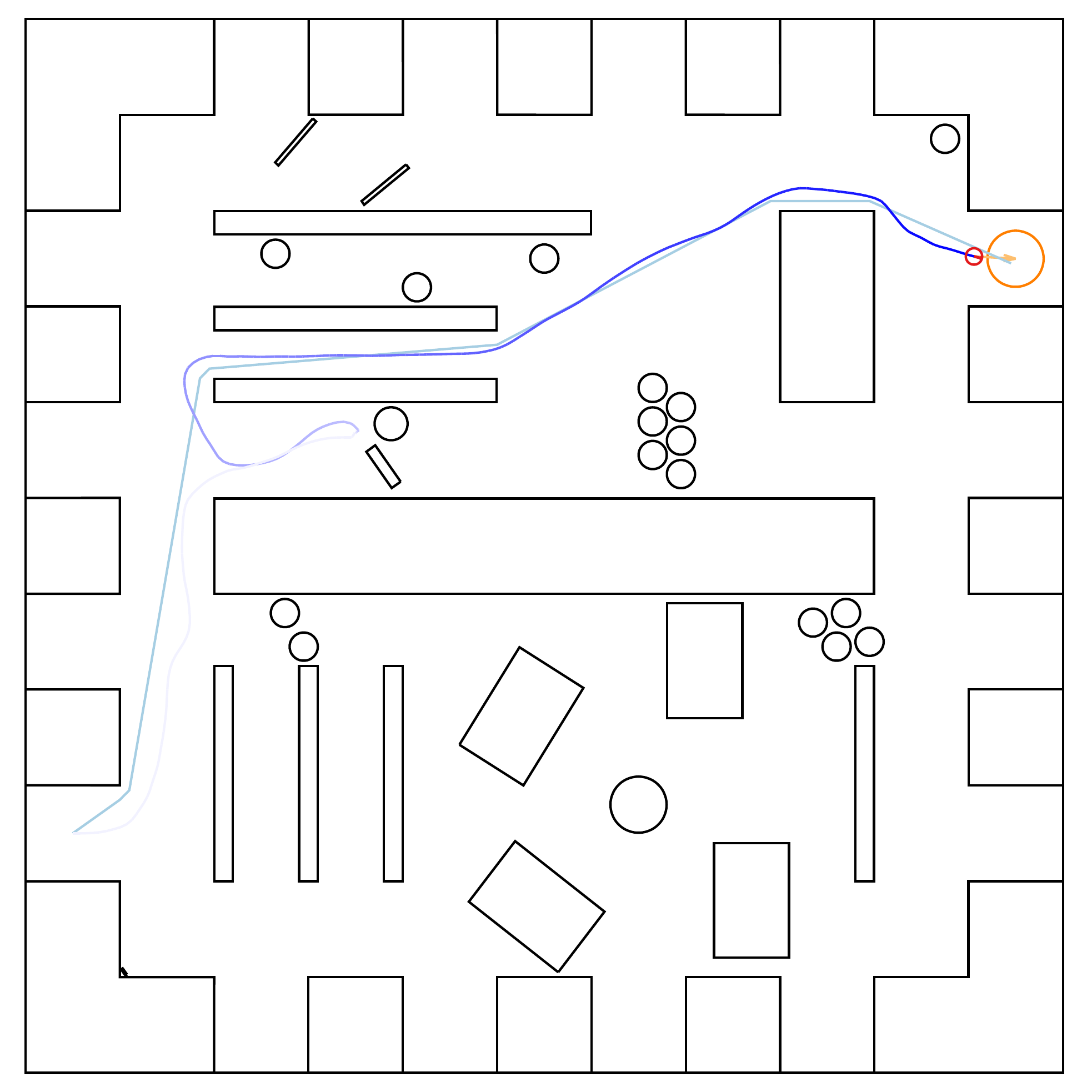}
		\caption{Hall with A-Star.}
		\label{fig:eval_hall_a_star}
	\end{subfigure}
	\caption{Exemplary visualization of agents moving in the known worlds.}
	\label{fig:eval_know_worlds}
\end{figure}

Despite of the decentralized system, in which an agent does not communicate with another agent, the agents learned an emergent behavior which is especially apparent in the multi-agent scenarios, shown in Fig.~\ref{fig:eval_know_worlds}{{a-c}}. In the environment ``swap'' both groups drive along their right side of the wall to allow the other group to pass by. In the environment ``constriction'' the two groups of agents merge like a zip or rather file in alternating sequence. In the ``intersection'' scenario the agents form a pattern in which the outer agents drive in a circle around the inner agents and the inner agents wait until the path is clear again.

Fig.~\ref{fig:eval_hall_a_star} shows the environment ``hall'' with a single agent. Hereby the path is planned with the A-Star algorithm to compare with the agent traveled distance. Without knowing the planned path of the A-Star the agent has chosen the same route, until the agent decided to drive into a dead end. This is because the distance and the orientation to goal rewards are higher in this moment. However, one has to highlight, that the agent learned to recover from the dead end and returns to the A-Star path, which was not shown e.g. in \cite{multi_agent_fan2020distributed}.

\subsection{Evaluation on Unknown Worlds}
\label{sec:unknown_worlds}
\begin{table*}[h!]
 \centering
 \caption{Results of evaluation in unknown worlds.}
 \label{tab:unknown_worlds}
 {\renewcommand{\arraystretch}{1.25}
 \begin{tabular}{|l|l|l|l||l|l|l|l|}
 \hline
 \multicolumn{4}{|c||}{\textbf{hall}} &  \multicolumn{4}{c|}{\textbf{roblab}} \\
 \hline
 \hline
 \textbf{Num. agents} & \textbf{Reached goal} & \textbf{Timeout} & \textbf{Collision} & \textbf{Num. agents} & \textbf{Reached goal} & \textbf{Timeout} & \textbf{Collision}\\
 \hline
 1  & \SI{74.00}{\percent} & \SI{23.70}{\percent} & \SI{2.30}{\percent} & 1  & \SI{51.70}{\percent} & \SI{3.40}{\percent} & \SI{44.90}{\percent} \\
 2  & \SI{71.30}{\percent} & \SI{27.00}{\percent} & \SI{1.70}{\percent} & 2  & \SI{51.50}{\percent} & \SI{3.05}{\percent} & \SI{45.45}{\percent} \\
 4  & \SI{71.00}{\percent} & \SI{27.35}{\percent} & \SI{1.65}{\percent} & 4  & \SI{51.85}{\percent} & \SI{3.43}{\percent} & \SI{44.72}{\percent} \\
 8  & \SI{71.73}{\percent} & \SI{26.71}{\percent} & \SI{1.56}{\percent} & 8  & \SI{50.45}{\percent} & \SI{4.21}{\percent} & \SI{45.34}{\percent}\\
 16 & \SI{72.86}{\percent} & \SI{25.77}{\percent} & \SI{1.37}{\percent} & 14 & \SI{48.46}{\percent} & \SI{5.51}{\percent} & \SI{46.03}{\percent}\\
 \hline
 \end{tabular}}
\end{table*}

In this section the agent is confronted with unseen environments. While the agent was trained only in the three environments ``tube'', ``room'', and ``four room'', the agent faced the problem with the two environments ``hall'' and ``roblab''. For both environment we evaluated the agent performance with increasing the number of agents. The number of the agents seems to have no effect on the agent's performance for both environments, as shown in Table~\ref{tab:unknown_worlds}.


The high timeout in the environment ``hall'' can be explained that the agent has not seen any dead end in the training and could not learn to recover from it. The obstacle in the world ``roblab'' are very small in comparison to the obstacles in the training environments. The convolution layers are not able to detect the small obstacles and the agent collided with them.

The trained agent with all environments mentioned in the section~\ref{sec:training} is set in the unknown environment multi with 24 agents. The environment ``multi'', shown in Fig.~\ref{fig:environment_multi}, is a mix of the multi-agent scenarios consisting of a combination of ``intersection'', ``bottleneck'' and some parts ``hall''. With the regular timeout of 500 steps the agent achieved a success rate of \SI{87.38}{\percent}, shown in Table~\ref{tab:unknown_worlds_2}.

\begin{table}[h!]
 \centering
 \caption{Results of evaluation in the unknown world ``multi'' with 500 and 1000 steps until timeout.}
 \label{tab:unknown_worlds_2}
 {\renewcommand{\arraystretch}{1.25}
 \begin{tabular}{|l|l|l|l|}
 \hline
 \textbf{Steps before timeout} & \textbf{Reached goal} & \textbf{Timeout} & \textbf{Collision} \\
 \hline
 500  & \SI{87.38}{\percent} & \SI{11.02}{\percent} & \SI{1.60}{\percent}\\
 1000 & \SI{93.32}{\percent} & \SI{4.37}{\percent}  & \SI{2.31}{\percent}\\
 \hline
 \end{tabular}}
\end{table}

By doubling the steps before a timeout, the agent reached its goal with \SI{93.32}{\percent}. A hard challenge for the agents is an intersection when multiple agents arrive simultaneously, but cannot pass the bottleneck at once. Hence some agents try to drive in other direction to reach the goal but get lost in a dead end.

We also tested the learned policy on a single physical robot which implies that the implemented simulation is aligned with the robot platform specification, as shown in following video \url{https://youtu.be/duXE8bnORIk}.
The Kobuki is chosen as modular robot platform with the NVIDIA Jetson Xavier as compute unit and the Hokuyo UST-20LX as 2D-laser scanner.
For obtaining the observation distance and orientation to goal the SLAM algorithm \cite{hector_slam_KohlbrecherMeyerStrykKlingaufFlexibleSlamSystem2011} is used.


\section{Conclusion}
\label{sec:conclusion}
In this paper we presented a decentral multi-robot navigation system which  has learned via DRL and is able to navigate through complex environments. It also can handle dead ends and reached its goal with \SI{99}{\percent} in the training environment. In complex unknown environments the learned policy still has a success rate of \SI{93}{\percent}. This all is achieved with the robust training algorithm PPO and the simple 2D-simulation which is designed for a fast and sample efficient training.

Despite the different environments the learned policy has some limitations where it cannot recover from large dead ends like in the environment multi is shown. In lots of other DRL problem the neuronal networks have recurrent elements to use the information about the past states, like e.g. LSTM \cite{lstm_hochreiter1997long}. Hereby the agent could remember visited locations and could handle larger dead ends. As well more environments diversity helps the policy to better generalized which can be achieved with a heuristically world generator where at every episode the environment is changed, as in \cite{procgen_cobbe2019leveraging}.

Frameworks like IMPALA \cite{impala2018} could be applied to our approach as they are designed for high parallel DRL training over multiple computers that are successfully used on very complex DRL problems \cite{alphastar_vinyals2019grandmaster}.



\clearpage
\acknowledgments{If a paper is accepted, the final camera-ready version will (and probably should) include acknowledgments. All acknowledgments go at the end of the paper, including thanks to reviewers who gave useful comments, to colleagues who contributed to the ideas, and to funding agencies and corporate sponsors that provided financial support.}


\bibliography{references}  

\end{document}